\documentclass[conf]{new-aiaa}
\usepackage[utf8]{inputenc}

\usepackage{svg}
\usepackage{booktabs}
\usepackage{cleveref}
\usepackage{graphicx}
\usepackage{amsmath}
\usepackage[version=4]{mhchem}
\usepackage{siunitx}
\usepackage{longtable,tabularx}
\usepackage{xspace}
\newcommand{\etal}{\emph{et al.}\xspace}
\usepackage{subcaption}
\title{Real-Time Wildfire Localization on the NASA Autonomous Modular Sensor using Deep Learning}

\author{Yajvan Ravan\footnote{OSTEM Internship Program, Langley Research Center}, Aref Malek\footnote{OSTEM Internship Program, Langley Research Center}, Chester V. Dolph\footnote{Aerospace Engineer, Aeronautics Systems Engineering Branch, AIAA Member.}, and Nikhil Behari\footnote{OSTEM Internship Program, Langley Research Center}}
\affil{NASA Langley Research Center, Hampton, VA, 23681}

\begin{document}

\maketitle

\begin{abstract}

    High-altitude, multi-spectral, aerial imagery is scarce and expensive to acquire, yet it is necessary for algorithmic advances and application of machine learning models to high-impact problems such as wildfire detection. We introduce a human-annotated dataset from the NASA Autonomous Modular Sensor (AMS) using 12-channel, medium to high altitude (3 - 50 km) aerial wildfire images similar to those used in current US wildfire missions. Our dataset combines spectral data from 12 different channels, including infrared (IR), short-wave IR (SWIR), and thermal. We take imagery from 20 wildfire missions and randomly sample small patches to generate over 4000 images with high variability, including occlusions by smoke/clouds, easily-confused false positives, and nighttime imagery. 

    We demonstrate results from a deep-learning model to automate the human-intensive process of fire perimeter determination. We train two deep neural networks, one for image classification and the other for pixel-level segmentation. The networks are combined into a unique real-time segmentation model to efficiently localize active wildfire on an incoming image feed. Our model achieves 96\% classification accuracy, 74\% Intersection-over-Union(IoU), and 84\% recall surpassing past methods, including models trained on satellite data and classical color-rule algorithms. By leveraging a multi-spectral dataset, our model is able to detect active wildfire at nighttime and behind clouds, while distinguishing between false positives. We find that data from the SWIR, IR, and thermal bands is the most important to distinguish fire perimeters. Our code and dataset can be found here: \hyperlink{https://github.com/nasa/Autonomous-Modular-Sensor-Wildfire-Segmentation/tree/main}{https://github.com/nasa/Autonomous-Modular-Sensor-Wildfire-Segmentation/tree/main} and \hyperlink{https://drive.google.com/drive/folders/1-u4vs9rqwkwgdeeeoUhftCxrfe\_4QPTn?usp=drive\_link}{https://drive.google.com/drive/folders/1-u4vs9rqwkwgdeeeoUhftCxrfe\_4QPTn?=usp=drive\_link}

\end{abstract}

\section{Introduction}
Wildfires are a pervasive and ever-growing problem across the world. With the advent of climate change, wildfires are becoming stronger, more frequent, and more unpredictable. From property damage to poor air quality, wildfires are the cause of many annual problems for those who live in areas prone to them. In 2021, wildfires generated a cost of \$15.5 billion dollars in damages and response \cite{acero}. Over 1 in 6 homes in the US is at risk of wildfires over the next 30 years \cite{acero}. Unfortunately, according to a joint workshop between NASA and the US Forest Service (USFS), numerous challenges are still present in wildfire management, including limited surveillance resources, accurate detection in cloudy conditions, limited data \& model fusion, and challenges with integrating new technologies amidst diverse stakeholders which requires simplicity \& interoperability \cite{acero}. Recent NASA work pursues other challenges presented in \cite{acero}. For example, \cite{nasa1} and \cite{nasa2} describe novel methods to estimate smoke emission and characterize smoke composition using emitted fire radiation.

Most current methods for wildfire imaging and localization involve human interpretation to label captured imagery and direct response teams. The National Infrared Operations (NIROPS) component of the National Interagency Fire Center (NIFC) manages wildfire mapping for incident management across the nation \cite{IkhanaPaper}. NIROPS relies on manned aircraft with an onboard infrared sensor to gather large-scale, full-fire aerial imagery at altitudes around 10,000 feet and transmit them to ground stations \cite{IkhanaPaper, NIFCWebsite}. Trained human firespotters then label these images with the location of the fire perimeter and heat sources, among other things, and this information is often used to direct response teams, oftentimes up to 12 hours after the original flight \cite{IkhanaPaper, NIFCWebsite}. 

The imaging aircraft employed today by the US Forest Service's (USFS) National Infrared Operations (NIROPS) \cite{IkhanaPaper, NIFCWebsite} generate medium-altitude (3-50km), wide-terrain, and wide-spectrum remote-sensing imagery. NASA’s Ikhana Western States Fire Missions (2007–2009) flew the AMS (Autonomous Modular Sensor) at ~12-13km and downlinked near-real-time fire products—hotspots, perimeters, and temperature-calibrated imagery—directly to incident command, demonstrating wide-area coverage with 3-50 meter ground sample distance (GSD) and reliable night operation \cite{IkhanaPaper,AMSPoster2014}. Today’s U.S. fire agencies continue to rely on this modality: the Forest Service’s NIROPS program operates high-altitude line-scanner systems (Phoenix/AMS) to map large incidents—most often at night—with geo-rectified TIFFs (file format) and shapefiles delivered before morning briefings; line-scanner scan volumes reach ~100k–400k acres/hour, enabling multiple large fires per sortie \cite{USFSFireImagingGuide2020,NIROPSInfraredBranch,PhoenixCapabilities}. State programs mirror this playbook: California’s Fire Integrated Real-Time Intelligence System (FIRIS, two King Air B200s) provides near-real-time perimeter and hotspot intelligence to California Department of Forestry and Fire Protection (CAL FIRE), California Governor’s Office of Emergency Services (CAL OES), and local agencies \cite{CalOESFIRIS,UCSDWIFIRE}, and Colorado’s Division of Fire Prevention and Control (DFPC) Multi-Mission Aircraft (PC-12, ~5.5-6km) streams IR/EO mapping to  Colorado Wildfire Information Management System (CO-WIMS) for incident use without congesting the tactical airspace \cite{ColoradoMMABrief2017,ColoradoMMAWIMS}. Beyond operations, NASA’s newer Compact Fire Infrared Sensing Testbed (c-FIRST) instrument flew January–March 2025 on the B-200 over the Palisades/Eaton fires to identify lingering hotspots, reinforcing the continued value of high-altitude thermal, infrared sensing for response and post-fire assessment \cite{NASAJPLCFirst2025,NASAScienceCFirst2025}.

Such a detection and response pipeline can be subject to expert error and could allow the fire to grow in the downtime between flight and response. Automation with computer vision has been explored in the domain of active wildfire detection. Some efforts have involved using algorithms with pre-determined thresholds for the intensities or ratios of various spectral bands \cite{SchroederPaper, MurphyPaper,KumarPaper}. However, this approach is strongly prone to false positives \cite{LandsatPaper}. More recent works have explored deep learning \cite{PDAMPaper,LandsatPaper,QuadTreePaper,SaliencyDetectionPaper}, but these focus on limited domains that are not widely used in current wildfire response missions, sticking to low-altitude imagery, satellite imagery, daytime imagery, RGB spectral data, or offline processing. As a result it is difficult to integrate these solutions into current wildfire operations.

This work is organized as follows. In Section \ref{sec:relatedwork}, we review past work towards solving this problem. In Section \ref{sec:methodology} we explain existing challenges in past work and describe our dataset and imagery. We then describe the neural networks that comprise our model, their training, evaluation, and combination. Finally, in Section \ref{sec:results} we explain the results of testing our model, benchmark it against previous work, depict examples of simulated real-time inference, and compare it with ablated versions of our model using less spectral data. With this work, we hope to demonstrate a proof-of-concept of an automated wildfire localization system that can be integrated alongside humans into the aforementioned response pipeline and spark future work in this domain to eventually automate this process entirely. Our main contributions are as follows:
\begin{enumerate}
    \item We create a \textbf{manually-annotated} dataset that mirrors modern-day wildfire observation missions using flights taken with the NASA Autonomous Modular Sensor (AMS) \cite{IkhanaPaper}
    \item We train classification and segmentation networks on our dataset and combine these into a two-tier wildfire localization model. We demonstrate our algorithm's ability to run in real-time on a simulated image feed.
    \item We show that the usage of IR imagery and hand-labelled data improves the performance, especially recall, of our model compared to past methods. 
\end{enumerate}

\section{Related Work}
\label{sec:relatedwork}
\subsection{Datasets}
Given the niche domain of this problem, very few datasets exist. This presents a sizeable obstacle to training deep learning models for this problem. A primary dataset is the Corsican Fire Database \cite{CorsicanDatasetPaper} which contains 500 RGB images and 100 multispectral images, of which only a few are aerial images of the type that would be taken from unmanned aerial vehicles (UAVs). Another dataset is the Visifire dataset \cite{visifiredatasetpaper} which contains seven videos of wildfires from an aerial point-of-view, but does not contain annotations. Pereira et. al created a custom dataset using Landsat-8 imagery and color-rule algorithms from \cite{MurphyPaper,KumarPaper,SchroederPaper} as ground-truth, along with a 1000 image manually annotated dataset consisting only of positive examples. The other works mentioned above used either the Corsican Fire Database or custom, unpublished datasets extracted from their own experiments or the Internet \cite{LandsatPaper}. 
Beyond Corsican and Visifire, several public resources target aerial or multi-sensor wildfire imagery. AIDER (Emergency Response) focuses on aerial disaster scene classification; it includes fire/smoke along with floods, collapsed buildings, and traffic accidents (320 fire/smoke images; 2,545 total including 1,200 “normal”), with an expanded AIDERv2 later released at scale \cite{kyrkou2019aider, kyrkou2020aider}. FLAME provides UAV-collected video and frames of prescribed pile burns, including RGB and multiple thermal palettes, with labeled frame sets for classification (39,375 train / 8,617 test) and 2,003 RGB/mask pairs for segmentation \cite{shamsoshoara2021flame}. FLAME-2 extends this idea to paired, time-synchronized RGB/LWIR from a DJI Mavic 2 Enterprise Advanced, releasing 53,451 paired frames for classification extracted from seven video pairs with four frame-level classes (flame/smoke combinations) \cite{chen2022flame2}. FASDD aggregates over 100,000 fire/smoke images across three sub-datasets (ground cameras, UAV, and remote-sensing), with bounding boxes and a georeferenced RS subset—useful for cross-domain detection and small-target scenarios \cite{wang2024fasdd}. This dataset is the most similar to ours. However, it contains only low-altitude airborne images and spaceborne images, without any data in between. Despite its abundance, satellite imagery differs distributionally from airborne remote-sensing data, and wildfire response missions favor airborne platforms because they can be directed and timed as needed. Numerous other datasets that are similar have emerged in recent years both with satellite imagery \cite{Tran2025Land8Fire} and relatively lower altitude UAV imagery \cite{Pesonen2025BorealForestFire,Kularatne2024FireManUAVRGBT, Hopkins2024FLAME3, Mowla2024UAVsFFDB, Mowla2024UAVsFFDB_DataInBrief, Zhang2025RGBT3M, Gomez2025GOES16SmokePlumes}.

\subsection{Image Segmentation}
Image classification and image segmentation are related but distinct: classification assigns a single label to an image, whereas semantic segmentation assigns a class label to each pixel \cite{SemanticSegmentationReviewPaper}. Progress over the past decade has been driven by deep learning and large benchmarks such as PASCAL VOC \cite{PascalVOCPaper} and MS COCO \cite{COCOPaper}, which span diverse object categories and scenes. Early deep-learning methods adapted CNN classifiers into fully convolutional networks (FCNs) for dense prediction \cite{FCNPaper}. Subsequent work improved context capture via spatial/pyramid pooling and dilated (atrous) convolutions \cite{PSPNetPaper,DilatedConvPaper}, and via encoder–decoder designs. A canonical example is U-Net, which uses skip connections to fuse fine and coarse features \cite{UNETPaper}, and has been effective across medical, aerial, and satellite imagery \cite{InriaAerialLabellingPaper,LandsatPaper}. Another widely used encoder–decoder is DeepLabv3+, which augments atrous spatial pyramid pooling with a lightweight decoder to sharpen boundaries, yielding strong results on PASCAL VOC and Cityscapes \cite{DeepLabV3PlusPaper,CityscapesPaper}.

More recently, transformer-based models have set the pace by modeling long-range dependencies with self-attention. Representative families include: (i) pure/ViT-based encoders such as SETR and Segmenter, which pair a transformer encoder with simple upsampling or a mask transformer decoder, achieving competitive accuracy on ADE20K and Cityscapes \cite{SETRPaper,SegmenterPaper,ADE20KPaper,CityscapesPaper}; (ii) hierarchical transformers such as SegFormer, which uses a multi-scale MiT encoder and an MLP decoder to balance accuracy and efficiency on ADE20K/Cityscapes \cite{SegFormerPaper}; and (iii) mask-classification transformers such as Mask2Former, which unify semantic, instance, and panoptic segmentation with a transformer decoder over mask embeddings, attaining state-of-the-art results on COCO, ADE20K, and Cityscapes \cite{Mask2FormerPaper,COCOPaper,ADE20KPaper,CityscapesPaper}. For instance segmentation, transformer approaches build on or surpass classical baselines like Mask R-CNN \cite{MaskRCNNPaper}. Together, these encoder–decoder CNNs and modern transformer architectures define today’s spectrum of high-performing segmentation methods.

\subsection{Color-Rule Algorithms}
The problem of wildfire detection is a simplified version of semantic image segmentation, where we wish to isolate which pixels correspond to fire in an image. The simplest solution to this problem relies on identifying fire by color, i.e. using color-space segmentation based on the available spectral channels \cite{UAVWildfireSurveyPaper}. For example, Murphy \etal \cite{MurphyPaper}, Schroeder \etal \cite{SchroederPaper}, and Kumar and Roy \cite{KumarPaper} developed color-value rule-based algorithms for the images captured by the LANDSAT-8 satellite \cite{landsatsatellitepaper}. These algorithms rely on techniques such as statistics in small neigbhorhoods around each pixel, reflectance ratios, and band saturation. An example rule from Schroeder \etal is shown in Figure \ref{fig:schroederalgorithm} where $\rho_i$ is the reflectance in channel $i$, and $R_{ij} = \rho_i/\rho_j$. However, these algorithms are strongly prone to false positives from objects with similar colors to fire and have limited robustness in real-world applications \cite{UAVWildfireSurveyPaper}.

\begin{figure}[!h]
    \begin{equation}
        \begin{split}
        \notag\{(R_{75} > 2.5) \text{ and } (\rho_7 - \rho_5 > 0.3)\text{ and } (\rho_7 > 0.5)\} \text{ OR } \\
  \{(\rho_6 > 0.8) \text{ and } (\rho_1 <  0.2) \text{ and } (\rho_5 > 0.4 \text{ or } \rho_7 < 0.1)\}
        \end{split}
    \end{equation}
  \label{eq:important}
  \caption{Algorithm for identifying unambiguous fire pixels according to Schroeder \etal \cite{SchroederPaper}}
  \label{fig:schroederalgorithm}
\end{figure}

\subsection{Deep Learning Approaches to Wildfire Detection}
More promising results come from the application of deep-learning methods to this problem. Lee \etal demonstrated classification networks based on a variety of aforementioned architectures \cite{UAVClassificationPaper}. The authors trained on around 20,000 images extracted from aerial videos from UAVs and were able to achieve up to 99\% accuracy of classification on fire and non-fire images \cite{UAVClassificationPaper}. Zhao \etal created the network architecture of FireNet, based on AlexNet, and demonstrated 98\% classification accuracy of fire and non-fire images \cite{SaliencyDetectionPaper}. The authors also augment their dataset using saliency-based localization methods to isolate regions of interest containing fire \cite{SaliencyDetectionPaper}. Park \etal explored the use of transfer learning with various architectures for multi-label classification of wildfire images \cite{object_detection_transfer}.

Much of the aforementioned work has focused on classification or isolation of regions of interest with fire. Frizzi \etal trained a network based on the VGG16 architecture to segment RGB images with fire and smoke classes and were able to achieve 72\% IoU for fire \cite{Fire&SmokeSegmentationPaper}. Lee \etal train a segmentation model based on the mobilenetv3 architecture to perform real-time (59 FPS) fire segmentation on RGB images in \cite{deeplabv3}. Behari \etal train a variety of network architectures for fire segmentation creating ground truth fire labels from Landsat-7 imagery by computing the difference normalized burn ratio (dNBR) index and taking pixels with a dNBR value of over 0.27 as active fire \cite{BehariPaper}. Pereira \etal generated ground truth masks of wildfire imagery taken with the Landsat-8 satellite \cite{landsatsatellitepaper} using color-rule algorithms from \cite{MurphyPaper,KumarPaper,SchroederPaper}. They then trained a U-Net based segmentation model with this data, experimenting further with using all 10 spectral channels vs. only 3, achieving up to 81\% accuracy on manually annotated data \cite{LandsatPaper}. Finally, Perrolas \etal the authors propose a search algorithm using the Quad-Tree method for fire localization \cite{QuadTreePaper}. Their algorithm recursively divides a larger image into quadrants if they contain fire as determined by their classification network based on SqueezeNet \cite{SqueezeNetPaper}. Then, once the patches are small enough, they perform segmentation using a U-Net based model. The authors propose a system that takes RGB images from an aerial view and transmits them to a processing unit to run their algorithm offline.

In addition to various segmentation methods, many works have explored variations of object detection techniques.  In \cite{yolofiredetectionpaper} the authors demonstrate the use of the YOLOV3 \cite{YOLOV3Paper} object detection algorithm to draw bounding boxes around regions of fire on RGB images with 83\% accuracy. In \cite{fireyolo}, Zhao \etal compared various object detection methods such as YOLOV3 and Faster R-CNN showing that they were better at detecting small fire targets but that they are prone to falsely detecting fire-like and smoke-like targets. In \cite{color-space-ann}, the authors combine color-space rules, texture rules, and deep learning to extract features and perform fire segmentation on image patches.

Recently transformers have become more common in the field of machine learning and computer vision, especially with the advent of vision transformers. For example \cite{transformerpaper}, the authors use a network architecture referred to as TransUNet to perform fire segmentation. They train their models using the Corsican Fire Database \cite{CorsicanDatasetPaper} achieving better performance compared to other deep learning methods.

\section{Methodology}
\label{sec:methodology}
In this work, we try to improve upon the existing literature in a few different ways. Firstly, we focus on the domain of high-altitude (3-50 km) remote sensing imagery, specifically using imagery obtained from the Ikhana flights \cite{IkhanaPaper}, and create a large manually annotated dataset. Airborne thermal and IR imaging has clear operational precedent and advantages for wildland fire intelligence. This data is similar to the medium-altitude, wide-terrain imagery generated by the imaging aircraft employed today by the NIROPS and the USFS \cite{IkhanaPaper, NIFCWebsite}, but such a human-labelled dataset is not publicly available. These images exhibit a different pattern than the data from aforementioned work such as \cite{Fire&SmokeSegmentationPaper, CorsicanDatasetPaper}, particularly in that active wildfire is a very small fraction of the image ($\sim$2\% on average). Compared to satellites, high-altitude airborne sensors occupy a useful “middle ground.” Satellites (e.g., MODIS/VIIRS) offer global persistence but are constrained by orbit schedules, cloud/smoke/terrain obscuration, and coarser native resolution (375 m for VIIRS I-band, ~1 km MODIS), which limits perimeter fidelity and small fire detection—especially under mixed-pixel conditions \cite{Giglio2016,Schroeder2014}. Compared to low-altitude UAVs, high-altitude line-scanners sacrifice centimeter-scale detail but deliver orders-of-magnitude higher scan rates and broad coverage not limited by Temporary Flight Restrictions (TFR); UAV/EO-IR “ball” systems excel for tactical overwatch and spot-fire confirmation but are endurance/FOV-limited and slower for whole-incident mapping \cite{USFSFireImagingGuide2020,Allison2016}. Finally, AMS-style systems also provide dual-gain thermal channels (high-gain sensitivity for subtle contrasts; low-gain headroom for ~800–1000 °C), letting a single overflight capture both faint residual heat and active flame fronts without saturation—precisely the regime of interest for our dataset’s high-altitude imagery \cite{AMSPoster2014,IkhanaPaper}.

Secondly, our paper makes use of all available spectral data including infrared (IR), thermal, and short-wave IR in addition to the visual spectrum (RGB). Previous works primarily utilize the visual spectrum (notable exception for \cite{LandsatPaper}) as data input primarily because such images are easily accessible, unlike imagery containing IR or SWIR \cite{QuadTreePaper, Fire&SmokeSegmentationPaper, transformerpaper}. However, this leaves room for improvements when vision is obscured by nighttime, clouds, fog, heavy fire smoke, among other obstacles or easily confused scenarios, such as red/orange objects or backgrounds. Despite these benefits, no large, manually labelled dataset containing IR imagery is publicly available to our knowledge. One possible reason is that IR imaging, like in NIROPS missions, must be done with specialized and expensive sensors. Furthermore, some systems may be of military origin, requiring export controls on the data.

Third, we demonstrate a simple real-time segmentation model. Our algorithm uses two neural networks, a classification network and a segmentation network. The former decides whether the current image contains fire and the latter localizes the fire if determined necessary by classifier. The usage of classification prevents useless segmentation, improving speed further. Our models run on patches of size 256 $\times$ 256 pixels. To simulate real-time imaging by an aircraft, we split our test images into a grid of fixed size patches, and input them into our model at the same rate that they were acquired according to flight logs. This patching method allows us to expand our dataset drastically as explained in Table \ref{dataset}. Furthermore, this ensures that our model is scalable (important for new wildfire technology according to \cite{acero}), continuing to operate even when mission parameters such as altitude may generate different outputs.

\subsection{Dataset Preparation}
\label{dataset}
\begin{table*}
  \centering
  \scalebox{0.99}{\begin{tabular}{c | c c c c }
    \toprule
    Dataset & GSD (m/pixel) & Altitude & Spectral Range & Images \\
    \midrule
    AMS (Ours) & 3-50m based on alt. & \textbf{Aerial (13.5km)} & \textbf{Blue - Thermal} & 5364\\
    Corsican Fire Database \text{\cite{CorsicanDatasetPaper}} & Ground & Ground & Blue - AMS Band 7 & 500  \\
    Visifire \text{\cite{visifiredatasetpaper}} & Ground  & Ground & Blue - Red & 2684  \\
    FASDD \text{\cite{wang2024fasdd}}& <1m or 10-30m
      & Ground/UAV/Satellite 
      & \textbf{Blue - Thermal} 
      & $>100{,}000$  \\
    Pereira Landsat \text{\cite{LandsatPaper}} & 30m & Satellite 705 km & Blue - SWIR & 150,000
    \text{\cite{landsatsatellitepaper}} \\
    Frizzi \etal \text{\cite{Fire&SmokeSegmentationPaper}} & Ground & Ground & Blue - Red & 8784 \\
    Perrolas \etal \text{\cite{QuadTreePaper}} & 1-5m & \textbf{Aerial} + Ground & Blue - Red & 820 + 500 \\
    FLAME\,2 \text{\cite{chen2022flame2}} & < 1m
      & UAV \(\sim\)60 m
      & \textbf{Blue - Thermal} 
      & 53{,}451 \\
    AIDER \text{\cite{kyrkou2019aider,kyrkou2020aider}} & < 1m 
      & UAV
      & Blue - Red
      & 2{,}545 \\
    \bottomrule
  \end{tabular}}
  \caption{Comparison of our dataset vs. other wildfire datasets. Public, human-annotated pixel-wise segmentation datasets in this table include AMS (ours), Corsican Fire Database, VisiFire (via the 2022 manual segmentation release), and FLAME-2. FLAME-2 provides paired RGB–LWIR imagery with masks for a 700-image subset, and Active Fire (Landsat-8) also includes a manually annotated subset. FASDD and AIDER do not provide pixel-wise masks. Frizzi \etal and Perrolas \etal don't provide human annotations. Our dataset is among those with the widest spectral range and is the largest aerial mid-to-high altitude manually-annotated set. }
  \label{tab:comparison}
\end{table*}

\begin{table}
  \centering
  \begin{tabular}{c | c c c}
    \toprule
    Band & Label & Wavelength, $\mu m$ & Features \\
    \midrule
    2 & Blue & 0.45-0.52 & Water \\
    3 & Green & 0.52-0.6 & Vegetation \\
    5 & Red & 0.63-0.69  & Heat Spots\\
    9 & SWIR II & 1.55-1.75 & Smoke/Heat\\
    10 & SWIR & 2.08-2.35 & Heat Spots\\
    11 & Infrared(IR) & 3.60-3.79 & Heat Spots\\
    12 & Thermal & 10.26-11.26 & Smoke/Heat\\
    \bottomrule
  \end{tabular}
  \caption{Visual Interpretation of the Frequency Bands captured by the AMS}
  \label{tab:importantbands}
\end{table}
Given the limited availability of human-annotated datasets in this domain, we decided to create our own to avoid the issue of false positives present in the aforementioned color-rule algorithms. A comparison of our dataset with other similar datasets is presented in Table \ref{tab:comparison}. It has the widest spectral range compared to previous datasets, contains the widest variety of difficult situations (occlusion by clouds, nighttime, and easily-mislabelled false positives). Furthermore, of the few datasets that contain aerial imagery, it is the only manually annotated one with over 1000 images.

\begin{figure}
  \centering
  \begin{subfigure}{0.3\linewidth}
    \includegraphics[width=\linewidth]{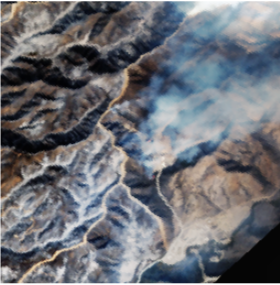}
    \caption{Visual Spectrum}
    \label{fig:datasetexample-a}
  \end{subfigure}
  \hfill
  \begin{subfigure}{0.3\linewidth}
    \includegraphics[width=\linewidth]{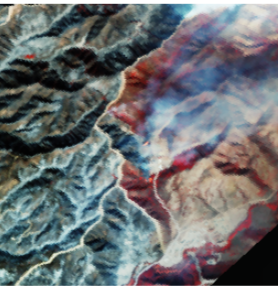}
    \caption{Color Infrared}
    \label{fig:datasetexample-b}
  \end{subfigure}
  \hfill
  \begin{subfigure}{0.3\linewidth}
    \includegraphics[width=\linewidth]{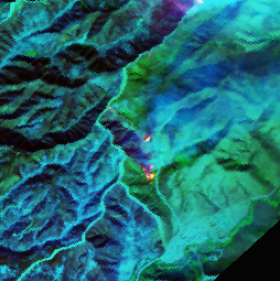}
    \caption{Fire Emphasis}
    \label{fig:datasetexample-c}
  \end{subfigure}
  \vfill
  \begin{subfigure}{0.3\linewidth}
    \includegraphics[width=\linewidth]{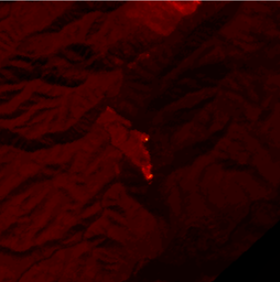}
    \caption{Thermal (12)}
    \label{fig:datasetexample-d}
  \end{subfigure}
  \hfill
  \begin{subfigure}{0.3\linewidth}
    \includegraphics[width=\linewidth]{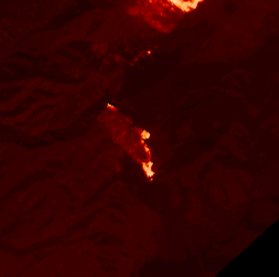}
    \caption{IR (11)}
    \label{fig:datasetexample-e}
  \end{subfigure}
  \hfill
  \begin{subfigure}{0.3\linewidth}
    \includegraphics[width=\linewidth]{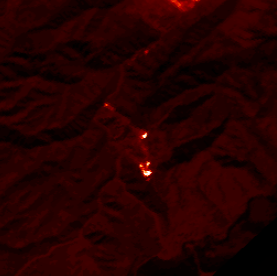}
    \caption{SWIR (10)}
    \label{fig:datasetexample-f}
  \end{subfigure}
  \hfill
  \caption{Example of Different AMS Spectral Bands From Our Dataset. \textbf{(a)} Bands Red/Green/Blue, \textbf{(b)} Bands 7/Red/Green, \textbf{(c)} Bands SWIR/7/Red, \textbf{(d)} Band 12 (Thermal), \textbf{(e)} Band 11 (IR), \textbf{(f)} Band 10 (SWIR). Active fire spots are clearly visible in c-f, but obscured in a-b.}
  \label{fig:datasetexample}
\end{figure}

We used 18 different observation missions completed by the NASA AMS sensor \cite{IkhanaPaper} ranging from 2006 to 2019 and in different times of day, smoke levels, and environments. Each image was captured over a target radius of typically 15-nm (equivalent to 1000 mi$^2$) at an altitude of 13.5km \cite{IkhanaPaper}. These images contained reflectance from 16 spectral bands. We chose to include only bands 1-12, since bands 13-16 are simply low-gain versions of bands 1-12 (with less temperature resolution). Of particular importance to our work were the bands outlined in Table \ref{tab:importantbands}. With this spectral data, different visualizations of a particular image are possible. An example of one image is shown in Figure \ref{fig:datasetexample}. Figure \ref{fig:datasetexample-a} shows the visual spectrum image using the red, green, and blue bands. Figure \ref{fig:datasetexample-b} uses a visualization from \cite{BehariPaper} called Color Infrared, using band 7, red, and green. This visualization highlights healthy vegetation in red \cite{BehariPaper}. Figure \ref{fig:datasetexample-c} uses another visualization from \cite{BehariPaper} called Fire Emphasis, using short-wave infrared (SWIR), band 7, and red. This visualization highlights active fires and hotspots in red \cite{BehariPaper}. Figures \ref{fig:datasetexample-d}, \ref{fig:datasetexample-e}, and \ref{fig:datasetexample-f} show the infrared (IR), thermal, and SWIR bands independently. In this particular example, the location or presence of fire is not clear with visual information alone, but with the presence of the heat-spotting bands (Thermal, IR, SWIR), it is clear.

\textbf{Preprocessing}. The AMS stores calibrated radiance values that are scaled and accompanied by extensive metadata including solar irradiance, solar zenith angle, temperature correction coefficients, emissivity, and calibration coefficients. The sensor records raw digital numbers (counts) in each band, which are linearly converted to radiance through comparison with blackbodies or ambient references. These radiance values are then scaled due to bit depth constraints and stored alongside scaling factors. For thermal channels (band 11, 12), radiance is calibrated using onboard blackbodies and onboard preprocessing yields brightness temperature values (in K) \cite{Ambrosia2011AMSWildfire,NASA_AMS_0700103_20061028,Gumley201x_MAS_IR, Hook2001MASTER}.

For non-thermal channels (bands 1-11), we normalize the calibrated radiance by the band-averaged solar irradiance. This approach approximates top-of-atmosphere (TOA) reflectance but omits solar zenith angle correction. This normalization serves multiple purposes: it standardizes channels relative to each other, accounts for slight variations in sensor positioning and lighting conditions, and preserves the visual distinction between daytime and nighttime imagery (nighttime images appear nearly black, as the eye would perceive them). The solar irradiance remains approximately constant across time and space for each band, providing near-constant normalization with minor corrections for lighting and sensor variation. We clip these normalized values between 0 and 1. While fire adds radiance to lower wavelength channels and can exceed unity, clipping does not significantly impact performance as the fire signal is captured primarily in the heat-sensitive bands (9-12). For thermal channels (12), we convert radiance to brightness temperature and clip values between 250 and 500 K, as temperatures exceeding this range provide diminishing utility for fire detection and normalization. Finally, similar to \cite{BehariPaper} we resample raw flight data to images with a consistent resolution of 10 m per pixel. This preprocessing pipeline is readily applicable to similar multi-spectral remote sensing platforms. We use 18 flights from the AMS for training and for testing, we generate another dataset from two flights using a similar remote-sensing instrument, the MODIS/ASTER airborne simulator (MASTER) \cite{Hook2001MASTER}, by choosing the 12 spectral bands that overlapped the most in frequency with the AMS.

\begin{figure}
  \centering
  \begin{subfigure}{\linewidth}
    \includegraphics[width=\linewidth]{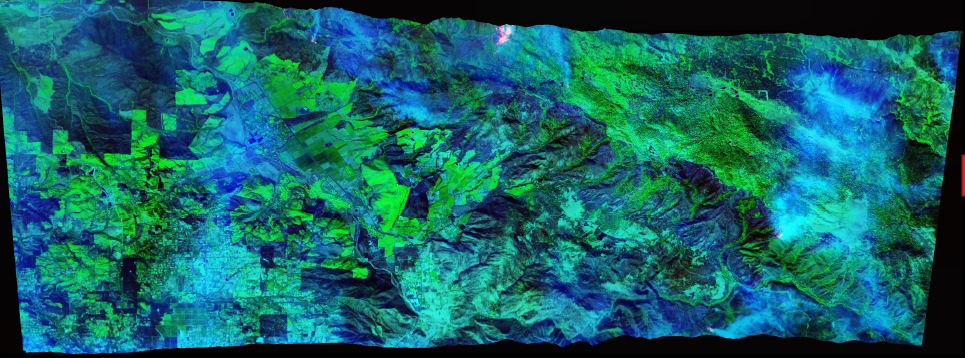}
    \label{fig:patchingexample-a}
    \captionsetup{font={small}}
    \caption{This image is from 1 of 11 California wildfires imaged on October 25, 2007}
  \end{subfigure}
  \vfill
  \begin{subfigure}{\linewidth}
  \centering
    \includegraphics[width=0.5\linewidth]{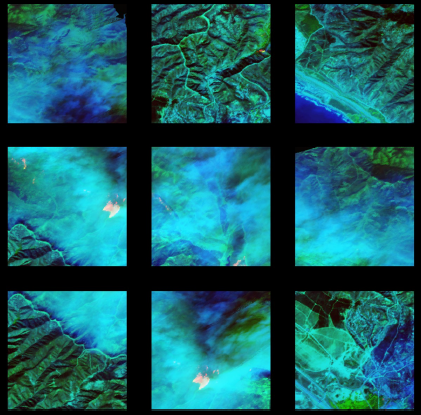}
    \captionsetup{font={small}}
    \caption{Randomly sampled patches drawn from the above. These patches differ greatly in smoke coverage, vegetation, human development, elevation, etc. This also minimizes the context forcing our model to learn fire detection with minimal information.}
    \label{fig:patchingexampleb}
  \end{subfigure}
  \caption{Demonstration of patching on AMS dataset}
  \label{fig:patchingexample}
\end{figure}

\textbf{Patching Augmentation} Due to the high cost of aerial missions, aerial imagery datasets tend to be small. However, due to the large area covered, we can generate 250-350 non-overlapping patches of size 256 $\times$ 256 pixels from each image. Different patches from the same image can look very different in terms of terrain as shown in Figure \ref{fig:patchingexampleb}. The small size of images also forces our model to learn the presence of fire with minimal context. The latter is particularly important as wildfire observation aircraft scan laterally and generate pieces of the image in series \cite{IkhanaPaper}, analogous to patching of the larger image. This resulted in 4259 patches for training (some with overlap) and 85 for testing (with no overlap). Our code is organized so that patches can be readily regenerated.

\textbf{Labelling} Each image was manually labeled by human inspection. To do so, all available spectral bands were used to mark, by hand, the location of active fire pixels on each image. An example of ground-truth label is shown in Figure \ref{fig:algorithmresults}. Only approximately 18\% of the patches contain more than 0.5\% active fire pixels.

\subsection{Networks}
\begin{figure}[!t]
    \centering
    \begin{subfigure}[t]{0.48\linewidth}
        \centering
        \includesvg[width=\linewidth]{figures/classification_architecture.svg}
        \caption{Classification Architecture: 3-layer encoder from 256×256×12 image to final binary output.
        Each layer uses Conv + Batch-Norm + Max-Pool.}
        \label{fig:classifiernetwork}
    \end{subfigure}
    \hfill
    \begin{subfigure}[t]{0.48\linewidth}
        \centering
        \includesvg[width=\linewidth]{figures/segmentation_architecture.svg}
        \caption{Segmentation Architecture: 2-layer encoder + 2-layer decoder from 256×256×12 input to 
        64×64×128 encoding and 256×256×1 output.}
        \label{fig:segmenternetwork}
    \end{subfigure}

    \vspace{1em}

    \includegraphics[width=0.9\linewidth]{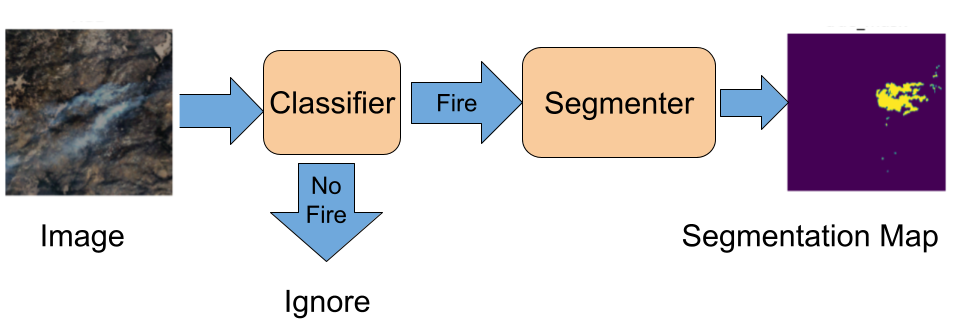}
    \caption{Inference model combining classification and segmentation networks. 
    Classification detects fire presence; segmentation localizes fire only when needed.}
    \label{fig:algorithmdiagram}
\end{figure}

The architecture of our classification network is shown in Figure \ref{fig:classifiernetwork}. Inspired by the encoder portion of U-Net \cite{UNETPaper}, it contains three layers of down-convolution (with ReLU activation) followed by batch normalization and max pooling. The final two layers are a global max pooling layer and a dense layer with sigmoid activation. 
The architecture of our segmentation network is shown in Figure \ref{fig:segmenternetwork}. Inspired by the segmentation network in \cite{QuadTreePaper}, it contains two layers of convolution and downsampling followed by two layers of deconvolution. We use skip connections between the encoder and decoder to maintain details from the high resolution images.

\subsection{Training and Evaluation}
The classification network was trained for 500 epochs using the binary cross-entropy loss function and the Adam optimizer \cite{adamppaper} using an initial learning rate of $7.5 * 10^{-4}$. We repeatedly sample positive examples to account for class imbalance. The segmentation network was trained using a weighted binary cross-entropy loss and the Adam optimizer \cite{adamppaper} with an initial learning rate of $3*10^{-4}$. Only positive examples were used to train the segmentation network, with a higher weight of 50 to balance the small fraction of the positive pixels (2\%). We evaluated our networks' performance based on accuracy, precision, recall, and IoU for segmentation. Because we intend for these networks to be used in real-time, we also evaluated the inference time for each network. 

\subsection{Two-Tier Model}
\label{sec:combination}
We combine these models into a real-time algorithm for wildfire localization. A schematic of our model is shown in Figure \ref{fig:algorithmdiagram}. For a given patch of size 256 $\times$ 256 pixels, the classifier network first determines if there is fire or not. If there is no fire in the patch (as is the case with 82\% of our dataset), then there is no need to run the larger segmentation network, saving computation time. If there is fire, then the segmentation network localizes it. On a static image, we split the image into a grid of patches and run inference on each one-by-one to simulate a real-time feed from an imaging aircraft.

We choose to create such a two-tier model for speed. The segmentation network contains 500 times the number of parameters in the classification network, drastically increasing its inference time (7$\times$, as shown in Table \ref{tab:testresults}). As aforementioned, the large flight paths and high-altitude of wildfire missions results in images that contain little fire, and this two-tier model prevents needless computation. When combined with the image patching approach in our dataset, this allows our model to operate faster on a real-time image feed. Furthermore, the partial interpretability offered by the two-tier system allows for cooperation with human experts as this model is integrated into wildfire missions. For example, the initial deployment may involve using only the classification network, allowing experts to ignore irrelevant portions of an image but still allowing for human-labelling of patches containing wildfire. Such integration can also be used to fine tune either network, while still saving time.

Although more complex compared to the color-rule algorithms, the usage of deep learning allows for greater accuracy, precision, and recall, as explained in Section \ref{sec:results}. However, we maintain as much simplicity as possible by keeping our networks shallow and using encoder-decoder segmentation techniques.   
\section{Results and Discussion}

\begin{table}[!h]
  \centering
  \scalebox{0.89}{
  \begin{tabular}{c | c c c c c c}
    \toprule
    Network & Acc. \%  & Prec. \%  & Recall \% & IoU \% & Inf. Time \\
    \midrule
    Our Classifier & 96.8 & 82.1 & 77.8 & & 1.19 ms \\
    Our Segmenter & 96.0 & 58.6 & 84.0 & 74.0  & 7.75 ms  \\
    Pereira \etal's segmenter & 95.5 & 53.2 & 74.8 & 47.5 & \\
    \bottomrule
  \end{tabular}}
  \caption{Results on the AMS test dataset, measuring \textbf{accuracy, precision, recall, IoU, and inference time}. Our model outperforms Pereira \etal's \cite{LandsatPaper} in all metrics, likely due to the use of human-annotated data and IR spectral data.}
  \label{tab:testresults}
\end{table}

\label{sec:results}
Although applying deep learning methods to high-altitude aerial imagery has not been thoroughly explored, Pereira \etal opened this exploration by constructing unique deep neural networks for wildfire segmentation on satellite imagery \cite{LandsatPaper}. As the closest approximation, we benchmarked our networks' performances against the segmentation network from \cite{LandsatPaper}, which we shall refer to as the Landsat network. The Landsat network was trained using Landsat-8 images that were \textbf{automatically} annotated by the aforementioned color-rule algorithms \cite{LandsatPaper}. According to the results in \cite{LandsatPaper}, this CNN has strictly greater recall, accuracy, and IoU compared to each of the color-rule algorithms from \cite{SchroederPaper, MurphyPaper, KumarPaper}. The Landsat network also only makes use of bands 10, 9, and 2, since according to the results in \cite{LandsatPaper}, such a network performs just as well as one using all available spectral data.  We measured the accuracy, precision, recall, and IoU (the ratio of the intersection between ground-truth and predicted fire areas and their union, with 1 indicating perfect prediction and 0 indicating no overlap).

The results of our models on our test set are shown in Table \ref{tab:testresults}. Results are averaged over 10 training seeds. The results show that our segmentation network outperforms the Landsat network, and therefore all of the color-rule algorithms as well. Our model surpasses it in precision (58.6\% vs 53.2\%), recall (84\% vs 74.8\%) and significantly in IoU (74\% vs 47.5\%).

We also chose to train our networks using only spectral bands 10, 9, and 2, which we refer to as the trimmed model, as this is directly comparable to the Landsat network. Results are shown in Section \ref{sec:ablation} and Table \ref{tab:classificationtestresults}, showing that our methods outperform the Landsat network and color-rule algorithms, as measured by IoU (75\% vs 47.5\%).

This latter comparison suggests that the use of human-annotated data improves network performance when tested on a human baseline, and also displays a distinct domain difference between satellite imagery and the airborne, high-altitude remote-sensing imagery captured by the AMS. The improvement in accuracy is largely due to a drop in the rate of false negatives, as can be seen in the difference in recall. In this domain, a high recall is arguably more important than high precision, as missed fire has significantly more consequences than a false alarm. Given that both our model and the Landsat network are based on similar UNet architecture, the key reason for performance lies in the datasets. As the color-rule algorithms are known to be inaccurate, the choice of manually-labelled data is strongly preferred, since automated labelling generates poor ground truth labels and therefore inaccurate network training. 

The full model is comparable with the trimmed model in most metrics, with a slight improvement in IoU between all of the trimmed versions and the full model, which indicates that the use of additional spectral, particularly the thermal and IR bands improves performance. Thus, these bands contain additonal information that can help distinguish potential errors in the short-wave IR band. Evidently, the improvement from the Landsat network is two-fold, both due to additional spectral information and due to the use of human-annotated data.

We present example results, shown in Figure \ref{fig:algorithmresults}. Figure \ref{fig:algorithmresults-a} shows a correctly segmented example. In this image, fire is occluded by clouds in the visual spectrum, and thus would not be caught by most of the aforementioned works. However, our model makes use of the IR band to localize the fire. Figure \ref{fig:algorithmresults-b} shows another correctly classified positive example. In this one, the active firespots are difficult to decipher among smoke and previously burned terrain, however, our model correctly localizes the fire, including pieces that are incredibly small and on the border.

\subsection{Real-Time Simulation}

\begin{figure*}[!h]
\centering
    \begin{subfigure}{\linewidth}
\centering
        \includegraphics[width=0.9\linewidth]{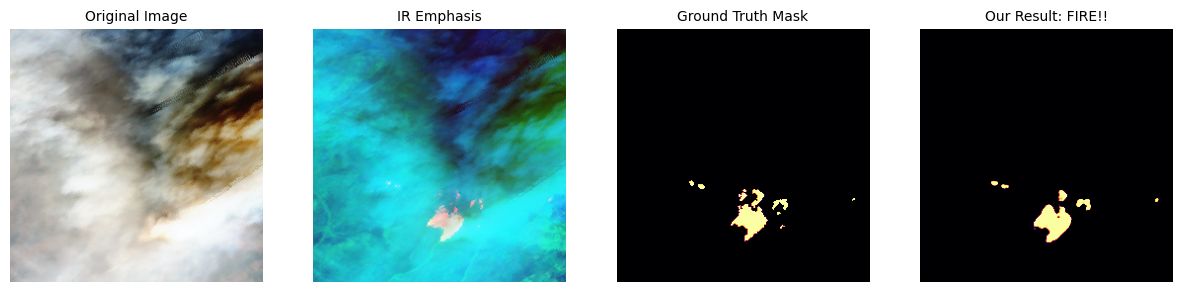}

    \captionsetup{font={small}}
   \caption{In the visual spectrum image, the fire is occluded by clouds, but is readily observable in the IR band.}
   \label{fig:algorithmresults-a}
    \end{subfigure}
    \vfill
\centering
    \begin{subfigure}{\linewidth}
\centering
        \includegraphics[width=0.9\linewidth]{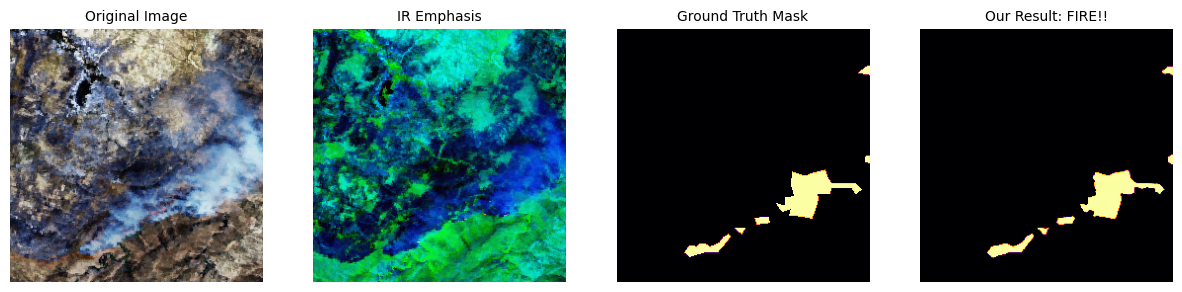}

    \captionsetup{font={small}}
   \caption{In the visual spectrum image, the fire is occluded by smoke and vegetation, but the burned area is readily observable in the IR band.}
   \label{fig:algorithmresults-b}
    \end{subfigure}
    \vfill
\centering
    \begin{subfigure}{0.9\linewidth}
\centering
        \includegraphics[width=\linewidth]{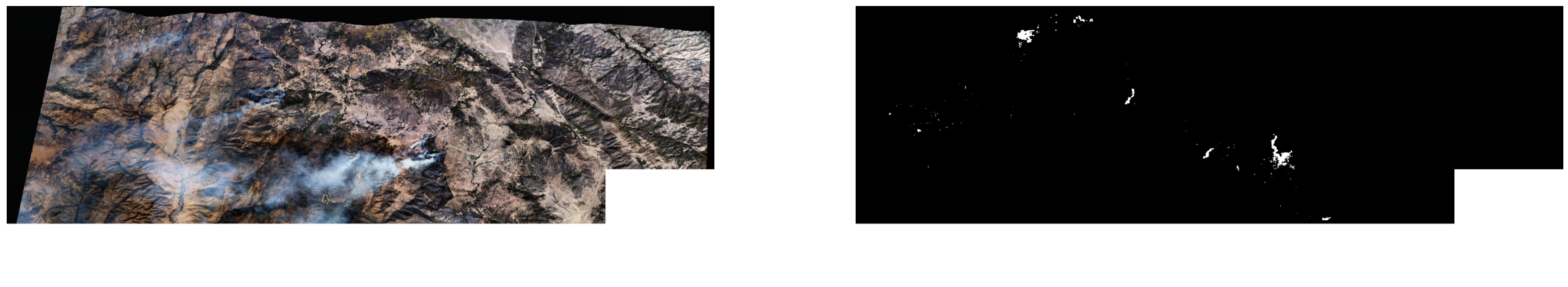}

    \captionsetup{font={small}}
   \caption{Example of real-time segmentation. The left image depicts the input image being fed in line-by-line (paused halfway through), and the right image depicts the output that our model produced, likewise. White is where fire was detected, and this correlates with simple visual determination based on smoke plumes.}
   \label{fig:algorithmresults-c}
    \end{subfigure}
    \caption{Example of Results. }
   \label{fig:algorithmresults}
\end{figure*}

The inference time of our networks is shown in Table \ref{tab:testresults}. The results indicate that both networks combined can run at a speed of over 100 fps on our NVIDIA T1200 laptop GPUs \footnote{Trade names or trademarks are used in the report for identification only and that this usage does not constitute an official endorsement, either expressed or implied, by NASA.}. Each flight with the AMS took roughly 5-10 minutes and generated 200-300 data patches of size 256 $\times$ 256 pixels, so our algorithm comfortably meets real-time processing requirements for the image feed considered.

In order to verify this, we validated our algorithm with a simulation of real-time aerial image capture. Images were split into a grid of patches of size 256 $\times$ 256 pixels. These patches were fed in one-by-one from left-to-right and top-to-bottom. Our model then runs inference, as described in Section \ref{sec:combination}, and generates 256 $\times$ 256 pixels masks which are either blank, if the classifier determines that there is no fire, or are segmentation output. These masks are laid side-by-side to create our final output. Results showed that the output was generated in negligible time compared to the input feed rate. An example of a partially-completed simulation is shown in Figure \ref{fig:algorithmresults-c} for clarification.

\subsection{Spectral Ablation}
\begin{table}[!h]
  \centering
  \begin{tabular}{c | c c c c c }
    \toprule
    Channels & Accuracy \%  & Precision \%  & Recall \% \\
    \midrule
    11, 9, 2 & 96.4 & 78.6 & 80.0  \\
    10, 9, 2 & \textbf{97.4} & 83.3 & \textbf{84.4}   \\
    9 & 97.1 & 88.9 & 74.4 \\
    10 & 95.4 & 85.3 & 71.1 \\
    11 & 94.4 & 66.9 & 62.2  \\
    12 & 95.8 & \textbf{97.5} & 63.3  \\
    1-8 & 89.8 & 30.0 & 4.4   \\
    5, 3, 2 & 89.5 & 12.2 & 7.8   \\
    
    \bottomrule
  \end{tabular}
  \caption{Classification Results Using Subsets of Spectral Data. High predictability was achieved with bands 10 and 11 only.}
  \label{tab:classificationtestresults}
\end{table}

\begin{table}
  \centering
  \begin{tabular}{c | c c c c c }
    \toprule
    Channels & Accuracy \%  & Precision \%  & Recall \% & IoU \%\\
    \midrule
    11, 9, 2 & 95.6 & 54.9 & \textbf{79.8} & 71.6   \\
    10, 9, 2 & 96.6 & 64.0 & 76.8 & 75.0   \\
    9 & 96.6 & \textbf{67.8} & 60.4 & 71.7 \\
    10 &\textbf{ 97.0 }& \textbf{67.8} & 76.6 & \textbf{76.4} \\
    11 & 96.4 & 57.7 & 66.2 & 71.5  \\
    12 & 95.8 & 58.4 & 63.4 & 69.5  \\
    1-8 & 94.7 & 21.5 & 1.1 & 47.8 \\
    5, 3, 2 & 94.6 & 18.4 & 1.8 & 48.1  \\
    \bottomrule
  \end{tabular}
  \caption{Segmentation Results Using Subset of Spectral Data. Performance is good but not comparable to full 12 band model}
  \label{tab:segmentationtestresults}
\end{table}

The contribution of various spectral bands to wildfire predictability was also tested by permuting over combinations shown in Table \ref{tab:classificationtestresults}. Specific subsets of spectral data were investigated, including the visual spectrum, fire emphasis, SWIR, pure IR, and pure thermal. Separate instances of classification and segmentation were trained using the same dataset and methods above (limited to the respective channel subset). The results are shown in Tables \ref{tab:classificationtestresults} and \ref{tab:segmentationtestresults}.

First, we note that the model using the RGB image or only the non-thermal bands results in very poor performance. The precision drops to at most 30\% for both classification and segmentation and the recall is below 10\%. Thus, we can conclude that much of the information needed to determine the presence of fire is contained in bands 9, 10, 11, and 12. Using either bands 9 or 10 achieves similar classification performance compared to the full model as noted by the recall and precision, while using either bands 11 or 12 achieves close but notably worse recall (for classification) or precision (for segmentation). 

 Thus, when limited model size is desirable (e.g. limited memory), one can use a classification network trained on this subset without sacrificing much performance. This also serves to inform that sensors should prioritize imaging in bands 9, 10, 11, and 12 or similar spectral ranges, as these contain the most important information for wildfire detection. Future work should also explore extracting information from these channels.

\label{sec:ablation}

\section{Conclusion}
\label{sec:conc}

The contribution of our work is three-fold. We explore the domain of high-altitude remote-sensing wildfire imagery to address the challenges in wildfire localization efforts. Firstly, we create and publish a novel human-annotated dataset derived from wildfires imaged by the NASA AMS sensor \cite{IkhanaPaper}. Second, we make use of 12 bands of spectral data to improve accuracy and recall, notably combining SWIR, IR, and thermal channels with the visual spectrum, an approach that, to our knowledge, has seen limited exploration to date in the context of wildfire segmentation. This implies that training data with complete spectral data is more useful than typical RGB images. Finally, we implement a real-time segmentation model for the purpose of scalability, ease-of-use, and efficiency. Future efforts in this area can explore methods for improving the performance of our algorithm. In order for such automation to be deployed in the field, one might desire to minimize false negatives, and hence high recall is desired. Another area to explore is the integration of this algorithm into existing pipelines. This will necessitate data preprocessing, and that step may not be negligible for real-time performance. Furthermore, humans are a necessity in these pipelines until algorithms like ours, become robust and reliable. Therefore, there is a need for exploring human-in-the-loop automation. Finally, as the field of computer vision progresses rapidly, many new techniques arise daily that have yet to be applied to this domain. As wildfires continue to be incredibly costly and deadly worldwide, this problem domain will remain a prime opportunity for automation using computer vision.

\section{Acknowledgements}
We wish to thank Anna Trujillo, Robert McSwain, Louis Glaab, Vince Ambrosia, and Marcus Johnson from NASA and Charles Kazimir from NIFC for their help. We would also like to thank the ACERO, STEREO, and TTT projects at NASA for funding this research.

\bibliography{sample}

\end{document}